\documentclass{article} %

\usepackage[table,dvipsnames]{xcolor}
\usepackage[most]{tcolorbox}
\usepackage{iclr2025_conference,times}

\usepackage{amsmath,amsfonts,bm}

\def\eqref#1{equation~\ref{#1}}

\def\1{\bm{1}}

\DeclareMathAlphabet{\mathsfit}{\encodingdefault}{\sfdefault}{m}{sl}
\SetMathAlphabet{\mathsfit}{bold}{\encodingdefault}{\sfdefault}{bx}{n}

\def\sR{{\mathbb{R}}}

\DeclareMathOperator*{\argmax}{arg\,max}
\DeclareMathOperator*{\argmin}{arg\,min}

\usepackage{hyperref}
\usepackage{url}
\usepackage{booktabs}
\usepackage{multicol}
\usepackage{multirow}
\usepackage{amssymb}
\usepackage{makecell}

\usepackage[noabbrev,capitalize]{cleveref}
\usepackage{enumitem}

\usepackage{theorem}

\usepackage[subtle,mathspacing=normal]{savetrees}

\theoremstyle{plain}
\newtheorem{theorem}{Theorem}[section]

\newtheorem{lemma}[theorem]{Lemma}

\theoremstyle{definition}

\theoremstyle{remark}

\title{Learning to Defer for Causal Discovery with Imperfect Experts}

\usepackage{fancyvrb}

\author{Oscar Clivio$^{1, }$\thanks{Work done while interning at ServiceNow Research, Mila and Université de Montréal.} \\
\And
Divyat Mahajan$^{2, 3}$ \\
\And
Perouz Taslakian$^{4,}$\thanks{Equal contribution.} \\
\And
Sara Magliacane$^{5,}$\footnotemark[2] \\
\And
Ioannis Mitliagkas$^{2, 3}$ \\
\And
Valentina Zantedeschi$^{4, 6,}$\thanks{Equal supervision.} \\
\And
Alexandre Drouin$^{2, 4, 6,}$\footnotemark[3] \\
\AND
\textnormal{$^1$ University of Oxford}\\ 
\And
\textnormal{$^2$ Mila – Quebec Artificial Intelligence Institute} \\
\And
\textnormal{$^3$ Université de Montréal} \\
\And
\textnormal{$^4$ ServiceNow Research} \\
\And
\textnormal{$^5$ University of Amsterdam}
\And
\textnormal{$^6$ Université Laval}
}

\iclrfinalcopy %

\allowdisplaybreaks
\begin{document}

\maketitle

\begin{abstract}
Integrating expert knowledge, e.g. from large language models, into causal discovery algorithms can be challenging when the knowledge is not guaranteed to be correct. Expert recommendations may contradict data-driven results, and their reliability can vary significantly depending on the domain or specific query. Existing methods based on soft constraints or inconsistencies in predicted causal relationships fail to account for these variations in expertise. To remedy this, we propose L2D-CD, a method for gauging the correctness of expert recommendations and optimally combining them with data-driven causal discovery results. By adapting learning-to-defer (L2D) algorithms for pairwise causal discovery (CD), we learn a deferral function that selects whether to rely on classical causal discovery methods using numerical data or expert recommendations based on textual meta-data. We evaluate L2D-CD on the canonical Tübingen pairs dataset and demonstrate its superior performance compared to both the causal discovery method and the expert used in isolation. Moreover, our approach identifies domains where the expert's performance is strong or weak. Finally, we outline a strategy for generalizing this approach to causal discovery on graphs with more than two variables, paving the way for further research in this area.
\end{abstract}

\section{Introduction}

Causal discovery is a fundamental task in artificial intelligence and data science, where the goal is to identify the unknown causal relationships between a set of random variables from their observations \citep{ spirtes2001cpas, pearl2009c, peters2017eocifala}. This typically involves inferring statistical dependencies in the data \citep{spirtes2001cpas}, leading to identification of the causal graph up to a Markov Equivalence Class (MEC), whose members all imply the same statistical dependencies~\citep{verma1990easocm, spirtes2001cpas}. Hence, we are not guaranteed a unique solution unless we make further distributional or functional assumptions to reduce the set of potential solutions~\citep{shimizu2006alngamfcd, peters2014cdwcanm}.

In addition to exploiting statistical dependencies, the use of large language models (LLMs) has recently gained significant attention in the causal discovery literature. This growing interest is largely driven by the promising performance of LLMs in various causal discovery studies \citep{willig2022cfmtc, jin2024cllmicfc}.
Typically, approaches query an LLM about the causal relationships between two or more variables, by prompting them with the variables' textual names, and optionally with additional information about the variables or an overall context \citep{abdulaal2023cmacgdtsmaddr, kiciman2023crallmoanffc, willig2023cpllmmtcbanc, long2023cdwlmaie, khatibi2024aalacdf, darvariu2024llmaepfcgd, long2024cllmbcg}. Other works attempt to use LLMs for conditional independence testing, which is then used for constraint-based causal discovery \citep{jin2024cacrilm, cohrs2024llmfcbcd}.

Thus, there has been a rising interest in combining knowledge-based causal discovery from LLMs with traditional statistical causal discovery \citep{choi2022lptlmatsp, long2024cllmbcg}. This research direction builds on significant work on combining more general expert knowledge with statistical causal discovery, notably to narrow down the set of potential solutions \citep{constantinou2023tiopkocsl}. Previous work has considered various types of expert knowledge: (i) required directed edges \citep{meek2013ciacewbk, decampos2007bnlausr, li2018bnslwsc} (ii) (partial)
orderings of the variables \citep{scheines1998ttpcbatcms, decampos2007bnlausr, andrews2020otcocditpolcwtbk, brouillard2022taiiicd}, (iii) ancestral constraints \citep{li2018bnslwsc, chen2016lbnwac}, and (iv) the negation of previous constraints \citep{meek2013ciacewbk, decampos2007bnlausr, chen2016lbnwac, li2018bnslwsc}. 

However, several challenges arise when using LLMs as experts for causal discovery. First, while LLMs are mostly effective at providing knowledge on ancestral and ordering relationships, they are far less reliable in determining  direct causal edges. This is because queries on causal relationships between two variables in natural language, such as ``does $X$ cause $Y$?'', typically fail to distinguish between direct and indirect effects, as this distinction depends on what other variables are observed~\citep{ban2023fqttcahllmfacdfd, kiciman2023crallmoanffc}. Hence, to integrate knowledge from LLMs with statistical causal discovery, we should use causal structures \citep{magliacane2017aci} or algorithms that allow for order or ancestral constraints \citep{chen2016lbnwac, ban2023fqttcahllmfacdfd, vashishtha2023ciulgd}. However, designing causal discovery algorithms with such constraints is generally more difficult, e.g. due to their non-decomposability \citep{chen2016lbnwac}. Further, LLMs might provide incorrect knowledge due to poor training data \citep{long2024cllmbcg}, lack of specialized knowledge or sensitivity to variations in words used in prompts to refer to causal relationships (e.g. ``cause'', ``influence'', etc.) \citep{darvariu2024llmaepfcgd}. This makes them ``imperfect experts'' \citep{long2023cdwlmaie}, necessitating methods that can accommodate incorrect knowledge. 

Towards this, a common approach is to incorporate background knowledge as ``soft constraints'' rather than hard constraints, often using Bayesian priors or initializations in score-based causal discovery methods \citep{choi2022lptlmatsp, dasilva2023hitlcdulcuag, darvariu2024llmaepfcgd}.
However, to the best of our knowledge, there is no consensus on how to select the appropriate prior and score, which are crucial for determining the final set of solutions and their correctness \citep{constantinou2023tiopkocsl, darvariu2024llmaepfcgd}. Alternatively, \citet{long2024cllmbcg} propose incorporating a model for the expert's correctness in the likelihood. However, this approach has limitations, such as using LLMs to query direct edges and assuming that imperfect experts make conditionally independent errors, an assumption that is often unrealistic as experts usually show systematic errors, e.g. they struggle on specific domains. Other methods involve LLM experts to estimate their own confidence \citep{zhang2023aeaatllmacmaad} or double check their own results \citep{ban2023fqttcahllmfacdfd}; but these are still vulnerable to the expert's mistakes. Finally, some works propose detecting inconsistencies caused by the incorrect expert knowledge \citep{decampos2007bnlausr, chen2023mpeicsltldpk}, however not all incorrect expert knowledge would cause such inconsistencies so this approach may not be sufficient to detect them. Crucially, none of these methods accounts for the features in each individual causal query, or more generally identifies which \emph{specific} knowledge returned by LLMs is correct and which is not, which is critical as LLMs might have better predictions on some domains such as common sense or insurance knowledge \citep{darvariu2024llmaepfcgd}, and struggle on others such as physics \citep{brown2020lmafsl} or specialized medical knowledge \citep{darvariu2024llmaepfcgd}.

\textbf{Contributions.} In contrast to previous work, we propose to directly \textit{learn and predict} which ancestral relationships returned by imperfect experts are correct, and which are not, informing the causal discovery process. To this aim, we build on the rich literature on \textit{learning to defer} (L2D) \citep{madras2018prifaabltd, mozannar2021cefltdtae} that consists in learning a deferral function: for any instance with features and an outcome, this function selects whether to predict the outcome using either a given machine learning model or an external black-box expert based on the features. While the deferral function and the model are most often learned jointly, recent works propose to learn the deferral function in a post-hoc manner, i.e. using already trained models \citep{narasimhan2022phefltdtae, mao2024tsltdwme}. We build on this methodology to learn whether to use an imperfect expert or a statistical causal discovery method to decide on the ancestry between two variables. Pairwise causal predictions can then be aggregated to form a topological order using techniques outlined in the extensive literature on ranking from noisy pairwise comparisons \citep{bradley1952raoibditmopc, feige1994cwni, rajkumar2014ascpoafrafpd, ren2021oscualbferfnc}.

Our key contributions are summarized below:
\begin{enumerate}
    \item We propose a learning-to-defer approach to integrate background knowledge in pairwise causal discovery; notably, we show how this reduces to a modified form of classification, allowing the use of any off-the-shelf classifier for this purpose. We call the resulting method \textbf{L2D-CD}.
    \item We show that this approach improves causal direction predictions in the canonical Tübingen pairs \citep{mooij2016dcfeuodmab}. The L2D-CD combination of an imperfect expert and a statistical causal discovery method outperforms each method alone and a simple deferral baseline that defers at random, showing the importance of \emph{learning} to defer.
    \item We describe how to extend this approach to graphs with $3$ variables or more, building on the literature on ranking from pairwise comparisons.
\end{enumerate}

\section{Methodology}

We consider the task of determining causal relationships between two causal variables, identified by their names ($u, v$). We denote the corresponding observational data with $N$ samples for these variables as $(x_u, x_v)$ where $x_u, x_v \in \sR^{N}$. Further, we assume access to metadata $C$ -- some textual context/description that provides extra information about the relationship between these variables. We represent the combined numerical data and textual description associated with the causal variables as $x= (C, u, v, x_u, x_v)$.

Let $y= \mathbb{I}_{u \rightarrow v}$, where $\mathbb{I}$ is the indicator function, denote a binary label for the causal relationship, i.e. $y = 1$ if $u$ causes $v$ and $y = 0$ otherwise. We assume access to the following predictors for causal relationships:
\begin{itemize}
    \item \textit{Expert Predictor.} Query an expert, typically an LLM, to obtain the causal relationship, where the expert uses only the textual description for prediction, i.e., $h_1(x)= \text{Expert}(C, u, v)$.
    \item \textit{Causal Discovery Methods.} Train a causal discovery method using a numerical/observational dataset to predict causal relationships, i.e., $h_0(x)= CD(x_u, x_v)$.
\end{itemize}

Our goal is to construct a predictor $h^{\star}(x)$ of $y$ that optimally combines the predictions from the expert $h_1(x)$ and the causal discovery method $h_0(x)$. 

\subsection{Background on Learning to Defer}

To combine multiple predictors, we use techniques from the literature on learning to defer (L2D)~\citep{madras2018prifaabltd, mozannar2021cefltdtae, mao2024tsltdwme}. Consider the task of binary prediction for input $x$ and labels $y$, with values in $\mathcal{X}$ and $\mathcal{Y}$ respectively, where $\mathcal{Y}$ is finite. Given the base predictor $h(x)=h_0(x)$ and $n_e$ expert predictors $\{ h_j(x) \}_{j=1}^{n_e}$, the goal is to learn a deferral function $r(x)= \argmax_{j \in [0, n_e]} r_j(x)$ that chooses between the different predictors for the sample $x$; e.g. $r(x) = 0$ means that $r$ chooses the base predictor $h$ and $r(x) = j$ means that $r$ chooses the $j$-th expert predictor $h_j$. Critically, only one of the base predictors and any expert predictor is chosen for a given instance $x$. As a result, the combined predictor $h^\star(x)$ can be constructed as $h^\star(x) = \sum_{j=0}^{n_e}\mathbb{I}_{r(x)=j}h_j(x)$.

In order to learn the deferral function, we use the loss objective from~\citet{mao2024tsltdwme}:

\begin{equation}
\label{eq:deferral-func-org}
L_{\text{def}}(h, r, x, y)= \mathbb{I}_{h(x) \neq y} \mathbb{I}_{r(x)=0} +  \sum_{j=1}^{n_e} c_j(x, y) \mathbb{I}_{r(x)=j},
\end{equation}
where $c_j(x,y) \in [0, 1]$ denotes the cost of predicting $y$ from instance $x$ associated with expert predictor $h(x)$, while the cost of the base predictor $h$ is the standard 0-1 loss. Essentially, $L_{\text{def}}$ is the loss incurred by the only base or expert predictor that $r$ chooses on instance $x$; indeed, only one of $\mathbb{I}_{r(x)=j}$ for $j = 0, 1, \dots, n_e$ is one and all others are zero. This leads to the following optimization problem,

\begin{equation}
\label{eq:opt-problem-def-org}
\argmin_{r} \mathbb{E}_{x,y} \big[ L_{\text{def}}(h, r, x, y) \big].
\end{equation}

L2D typically learns both $h$ and $r$ jointly \citep{madras2018prifaabltd}; however in our setting both causal direction predictors are pre-fitted, so we choose the version of \citet{mao2024tsltdwme} where $r$ is learned \textit{after} $h$ has been learned. We refer to this setup as \textit{post-hoc L2D} and to the two steps of learning $h$ then $r$ as \textit{two-stage L2D}. In the rest of the manuscript, unless specified otherwise, L2D refers to post-hoc L2D. Since the above loss is not differentiable for learning the deferral function $r$, the literature typically considers surrogate losses; for example \citet{mao2024tsltdwme} use the following surrogate loss:
\begin{equation}    
\begin{aligned}
\label{eq:deferral-func-surr}
L^h_{\text{surr}}(r, x, y) = \mathbb{I}_{h(x) = y} l_2(r, x, 0) +  \sum_{j=1}^{n_e} {\bar c}_j(x, y) l_2(r, x, j)
\end{aligned}
\end{equation}
where ${\bar c}_j(x, y)= 1 - c_j(x, y)$ and $l_2(r,x,y)$ is a surrogate loss using soft assignment functions $r_j(x)$ for multi-class classification problem with classifier $r(x)$. If we set $l_2$ to the logistic classification loss, we obtain the following:
\begin{equation}
    L^h_{\text{surr}}(r, x, y)= -\mathbb{I}_{h(x) = y} \log \frac{1}{1 + \sum_{k=1}^{n_e} e^{-r_k(x)} } - \sum_{j=1}^{n_e}  {\bar c}_j(x, y) \log \frac{e^{-r_j(x)}}{1 + \sum_{k=1}^{n_e} e^{-r_k(x)} }.
\end{equation}

Thus, L2D can be instantiated to our setting as $n_e = 1$, $x = (C,u,v,x_u,x_v)$, $y = \mathbb{I}_{u \rightarrow v}$, $h_1(x) = \text{Expert}(C, u, v)$, $h(x) = h_0(x)= CD(x_u, x_v)$. It is natural to use $c_1(x,y) = \mathbb{I}_{h_1(x) \neq y}$; however, this setup is excluded by \citet{mao2024tsltdwme}, since it violates an assumption used to prove desirable theoretical properties of the L2D surrogate loss. Nevertheless, as we will show next, these properties still extend to $c_1(x,y) = \mathbb{I}_{h_1(x) \neq y}$ when $y$ is binary, and this choice of $c_1(x,y)$ allows us to reduce L2D to standard classification when $n_e = 1$.

\subsection{Consistency bounds for the L2D loss with a 0-1 cost function and binary labels}

L2D typically uses a surrogate loss; however does minimizing it actually minimize the original loss? To assess this, we employ the concept of $\mathcal{H}$-consistency bounds \citep{awasthi2022hcbfslm, mao2024tsltdwme}. Let $\mathcal{H}$ be a hypothesis class, $\ell$ be a non-negative function over $(h,x,y)$ where $h \in \mathcal{H}$ is the predictor, $x$ the features and $y$ the label. Let $\mathcal{E}_\ell(h) := \mathbb{E}_{x,y}\left[\ell(h,x,y)\right]$ and    $\mathcal{E}_\ell^\star(\mathcal{H}) := \min_{h \in \mathcal{H}} \mathcal{E}_\ell(h)$. Further, denote $\mathcal{M}_{\ell}(\mathcal{H}) := \mathcal{E}_\ell^\star(\mathcal{H}) - \mathbb{E}_x\left[\inf_{h \in \mathcal{H}} \mathbb{E}_{y|x}[\ell(h,x,y)]\right]$ the minimizability gap of $\mathcal{H}$ and $\ell$, which is non-negative and vanishes when $\mathcal{E}_\ell^\star(\mathcal{H})$ coincides with the Bayes error of $\ell$. Then, an $\mathcal{H}$-consistency bound of a surrogate loss $\ell_s$ with respect to an original function $\ell_o$ is a bound of the form
\begin{align*}
    \forall h \in \mathcal{H}, \ \mathcal{E}_{\ell_o}(h) - \mathcal{E}^\star_{\ell_o}(\mathcal{H}) + \mathcal{M}_{\ell_o}(\mathcal{H}) \leq \Gamma\left(\mathcal{E}_{\ell_s}(h) - \mathcal{E}^\star_{\ell_s}(\mathcal{H}) + \mathcal{M}_{\ell_s}(\mathcal{H})\right),
\end{align*}
where $\Gamma$ is a non-decreasing concave function such that $\Gamma(0) = 0$. Such a bound is desirable as it implies Bayes-consistency of $\ell_s$ w.r.t. $\ell_o$, while also quantifying how improvements in $\mathcal{E}_{\ell_s}(h)$ translate to improvements in $\mathcal{E}_{\ell_o}(h)$ for $h \in \mathcal{H}$. From now on, we use ``consistency bound'' to refer to an $\mathcal{H}$-consistency bound without specifying the hypothesis class $\mathcal{H}$.

Theorem 6 of \citet{mao2024tsltdwme} shows that when the surrogate loss used to learn the predictor $h$ and the surrogate loss for the deferral function $r$ have consistency bounds with respect to the multi-class 0-1 loss, then the L2D surrogate loss $L_\text{surr}$ has a generalized form of consistency bound with respect to the original L2D loss $L_\text{def}$ in Equation \ref{eq:deferral-func-org}. However, this result relies on the assumption that
\begin{align*}
    \underline{c}_j \leq \bar{c}_j(x,y) \leq \bar{c}_j, \ \ \forall j,x,y
\end{align*}
for some $\underline{c}_j > 0$ and $\bar{c}_j \leq 1$, which does not apply when $c_j(x,y) = \mathbb{I}_{h_j(x) \neq y}$ for which $\forall j, \ \  c_j(x,y) = 0$ for some $x,y$. However, as we show next, the L2D loss does have such a consistency bound in this case, if one further assumes a binary $y$. We provide the proof in Appendix~\ref{supp:proof}.

\begin{lemma} \label{lem:bound}
    Assume that $y$ is binary and $\forall j = 1, \dots, n_e, \ \ c_j(x,y) = \mathbb{I}_{h_j(x) \neq y}$. Denote $\mathcal{H}$ and $\mathcal{R}$ hypothesis classes for the base predictor $h$ and the deferral function $r$, respectively. Assume that $l_1$ has a $\mathcal{H}$-consistency bound w.r.t. the binary 0-1 loss with concave function $\Gamma_1$, and $l_2$ has a $\mathcal{R}$-consistency bound w.r.t. the multiclass (with $n_e+1$ classes) 0-1 loss with concave function $\Gamma_2$. Then, for all $h \in \mathcal{H}$ and $r \in \mathcal{R}$,
    \begin{align*}
        &\mathcal{E}_{L_{\text{def}}}(h, r) - \mathcal{E}^*_{L_{\text{def}}}(\mathcal{H}, \mathcal{R}) + \mathcal{M}_{L_{\text{def}}}(\mathcal{H}, \mathcal{R})\\
        &\leq \Gamma_1(\mathcal{E}_{\ell_1}(h, r) - \mathcal{E}^*_{\ell_1}(h) + \mathcal{M}_{\ell_1}(\mathcal{H})) + n_e \Gamma_2\left(\mathcal{E}_{L^h_{\text{surr}}}(r) - \mathcal{E}^*_{L^h_{\text{surr}}}(r) + \mathcal{M}_{L^h_{\text{surr}}}(\mathcal{R})\right)
    \end{align*}
    where the $n_e$ factor can be removed when $\Gamma_2$ is linear. In particular, if $\mathcal{H}$ is a singleton $\{h_0\}$, then this reduces to a $\mathcal{R}$-consistency bound of $L^{h_0}_{\text{surr}}$ w.r.t. $L^{h_0}_{\text{def}}(r,x,y) := L_{\text{def}}(h_0,r,x,y)$.
\end{lemma}
Notably, Lemma \ref{lem:bound} will ensure that one can minimize the surrogate loss in Equation \ref{eq:deferral-func-surr} with guarantees on the combined predictor $h^\star(x)$ in our pairwise causal discovery setting

\subsection{Post-hoc L2D with a 0-1 cost function and a single expert can be reduced \\to standard classification}

We now show that post-hoc L2D applied to our setting can be reduced to binary classification. Indeed, assume $c_j(x,y) = \mathbb{I}_{h_1(x) \neq y}$ and let
\begin{align*}
    \mathcal{D}_\circ = \{(x,y) \ \ \ | \ \ \ \mathbb{I}_{h(x) \neq y} = \mathbb{I}_{h_j(x) \neq y} \ \ \forall j = 1, \dots, n_e\}
\end{align*}
which is the set of pairs of features and labels on which all predictors are equally correct or wrong. Then it turns out that for any $(x,y) \in \mathcal{D}_\circ$, $L_\text{def}(h,r,x,y) = \mathbb{I}_{h(x) \neq y}$ does not depend on $r$. In particular, noting  $\mathcal{D}^\complement_\circ$ the complement of $\mathcal{D}_\circ$ this leads to
\begin{align*}
\mathbb{E}_{x,y}\big[ L_{\text{def}}(h, r, x, y) \big] &= p(\mathcal{D}^\complement_\circ)\mathbb{E}_{x,y}\big[ L_{\text{def}}(h, r, x, y) \big| (x,y) \notin \mathcal{D}_\circ \big] + p(\mathcal{D}_\circ)\mathbb{E}_{x,y}\big[ \mathbb{I}_{h(x) \neq y} \big| (x,y) \in \mathcal{D}_\circ \big]
\end{align*}
Thus, we target the alternative loss $\mathbb{E}_{x,y}\big[ L_{\text{def}}(h, r, x, y) \big| (x,y) \notin \mathcal{D}_\circ \big]$ instead of the original loss $\mathbb{E}_{x,y}\big[ L_{\text{def}}(h, r, x, y) \big]$. Further, when $n_e = 1$, for any $(x,y) \notin \mathcal{D}_\circ$, then $\mathbb{I}_{h(x) \neq y} = \mathbb{I}_{h_1(x) = y}$. Thus,
\begin{align*}
    L_{\text{def}}(h, r, x, y) = \mathbb{I}_{h_1(x) = y}\mathbb{I}_{r(x) = 0} + \mathbb{I}_{h_1(x) \neq y}\mathbb{I}_{r(x) = 1} = \mathbb{I}_{r(x) \neq \mathbb{I}_{h_1(x) = y}}
\end{align*}
is the 0-1 classification loss for features $x$ and label $\mathbb{I}_{h_1(x) = y}$, which is the indicator that $h_1(x)$ returns the correct prediction $y$. This amounts to learning to predict when the expert is correct or, equivalently, when the base predictor is wrong. As a result, in this setting, post-hoc L2D reduces to binary classification over samples not in $\mathcal{D}_\circ$. This allows us to use any off-the-shelf classification method for training, while directly optimizing the expected surrogate loss without removing samples in $\mathcal{D}_\circ$ would generally only be feasible through automatic differentiation, as done in the implementation of \citep{mao2024tsltdwme}.

\subsection{Proposed Approach: L2D-CD}
\label{sec:training_procedure}

Based on the above discussion, our training procedure to obtain a fitted deferral function $\hat{r}$ from a set of $m$ examples $(x_i,y_i)_{i=1}^m$  
when $n_e = 1$ is thus to:
\begin{enumerate}
    \item Compute the set $S = \{i \ | \ \mathbb{I}_{h(x_i) \neq y_i} \neq \mathbb{I}_{h_1(x_i) \neq y_i}\}$. Note that when $y$ is binary, we generally have $\mathcal{D}_\circ = \{(x,y) \  | \  x \in \mathcal{X}_\circ\}$ where $\mathcal{X}_\circ := \{x \ \ \ | \ \ \ h(x) = h_j(x) \ \ \forall j = 1, \dots, n_e\}$, so here $S$ can also be computed as $S = \{i \ | \ h(x_i) \neq h_1(x_i)\}$.
    \item Obtain $\hat{r}$ by fitting any binary classification method to the set $(x_i,y'_i)_{i \in S}$ where $y'_i = \mathbb{I}_{h_1(x_i) = y_i}$.
\end{enumerate}

For pairwise causal discovery, we apply this procedure to $x = (C,u,v,x_u,x_v)$, $y = \mathbb{I}_{u \rightarrow v}$, $h_1(x) = \text{Expert}(C, u, v)$, $h(x) = h_0(x)= CD(x_u, x_v)$. We call the resulting method \textbf{L2D-CD}. Thus, intuitively L2D-CD learns which causal discovery predictor, statistical-based method or metadata-based expert, returns the correct causal direction when the two predictors differ in their predictions.

\section{Results on Tübingen datasets}
\label{sec:tuebingen}

We now apply the approach from Section \ref{sec:training_procedure} to the Tübingen pairs \citep{mooij2016dcfeuodmab}, a canonical benchmark for pairwise causal discovery.

\textbf{Division of Tübingen pairs by domain:} In order to design synthetic experts having strengths in certain domains and, potentially, weaknesses in others,
we manually assign a domain to each Tübingen pair. After excluding multivariate pairs (52-55, 71, 105), Tübingen pairs are each assigned to five different domains: Climate/Environment, Economics/Finance, Biology, Medicine, Physics. Further, a training set and a testing set are formed by stratified sampling w.r.t. the domains at 50/50 proportions. Table \ref{tab:pairs} in Appendix \ref{app:pairs} details the pairs present in each domain as well as in each of the training and testing sets. %

\textbf{Causal discovery methods:} We consider three causal discovery (CD) methods that can be applied on cause-effect pairs: LiNGAM (more specifically,  Direct-LiNGAM \citep{shimizu2011dadmflalngsem}), RECI \citep{blobaum2018ceibcre} and bQCD \citep{tagasovska2020dcfeuqbqcd}. Notably, LiNGAM assumes linear structural equations with non-Gaussian noise, while RECI and bQCD rely on a postulate of independence of causal mechanisms \citep{peters2017eocifala}; this means that they may also be misspecified, thus ``imperfect'', if such assumptions are violated, further strengthening the motivation of combining them with experts.

\textbf{Synthetic experts:} We design synthetic experts having a pre-defined probability $p_d \in [0,1]$ of returning the correct answer for each domain $d \in \{$Climate/Environment, Economics/Finance, Biology, Medicine, Physics$\}$, i.e. for each pair $i \in d$, the expert returns the correct causal direction with probability $p_d$ and the incorrect direction with probability $1 - p_d$. We consider two types of experts:

\begin{itemize}
\item $\epsilon$-experts : for a given $\epsilon \in (0,0.5)$, define:
    \begin{equation}
        p_{\text{Biology}} = p_{\text{Economics/Finance}} = p_{\text{Physics}} = 1 - \epsilon \ \text{ and } \ p_{\text{Climate/Environment}} = p_{\text{Medicine}} = \epsilon \label{eq:epsilon_experts}
    \end{equation}
    Notably, the expert is considered as ``good'' on domains $d$ where $p_d = 1 - \epsilon > 0.5$, and ``bad'' on domains $d$ where $p_d = \epsilon < 0.5$. Domains in Equation  \ref{eq:epsilon_experts} are chosen in order to assure a roughly equal share between pairs in a good domain (54) and pairs in a bad domain (48). In the following, we refer to these experts as ``$\epsilon=\dots$'' where ``$\dots$'' refers to the selected value of $\epsilon$. We considered $\epsilon = 0.05$, $\epsilon = 0.1$, and $\epsilon = 0.2$.
    
\item $p$-experts : here we consider deterministic experts, i.e. $p_d \in \{0,1\} \ \ \forall d$. However, unlike the previous experts, we change values of $p_d$ by domain $d$. To maintain consistency with the previous $\epsilon$-experts where three domains are good and two are bad,  we select assignments of $p_d$'s such that three $p_d$'s are equal to 1 and two are equal to 0. In the following, we refer to these experts as ``ABC'' were A, B, C refer to the initials of the domains $d$ where $p_d = 1$
\end{itemize}

\textbf{LLM experts} : Finally, we consider experts implemented using OpenAI GPT models \citep{openai2024gosc}. For every Tübingen pair $i$ with original description $D_i$, we manually remove ground-truth mentions in $D_i$ to obtain a ground-truth-free description $D'_i$ (see \cref{fig:text_description} for an example); and prompt the OpenAI model as follows :

\begin{itemize}
\item System part of the prompt : \emph{You will be given a text describing two columns in a dataset. The text will be delimited by backticks as in a code block. The first column is also referred to as ``x'' and the second column as ``y''. Based on the text description between backticks, is it more likely that 1) x causes y, or that 2) y causes x? Please choose one and only one of these two options.}

\item User part of the prompt : the text description $D'_i$ between two sets of three backticks and two spaces, as in a code block. An example of such $D'_i$ is given in Figure \ref{fig:text_description}.

\end{itemize}

\begin{figure}
    \centering
    \begin{tcolorbox}[colback=gray!10, colframe=black, sharp corners=south, width=0.9\textwidth]
        \textbf{Information for pairs0008:}\\[-2mm]

        \texttt{https://archive.ics.uci.edu/ml/datasets/Abalone}\\[-2mm]

        \begin{enumerate}[leftmargin=13pt]
            \item \textbf{Title of Database:} Abalone data

            \item \textbf{Sources:}
                \begin{itemize}
                    \item[a)] \textbf{Original owners of database:}
                    Marine Resources Division  
                    Marine Research Laboratories - Taroona  
                    Department of Primary Industry and Fisheries, Tasmania  
                    GPO Box 619F, Hobart, Tasmania 7001, Australia  
                    \textit{(contact: Warwick Nash, +61 02 277277, wnash@dpi.tas.gov.au)}
                    \item[b)] \textbf{Donor of database:}
                    Sam Waugh (\texttt{Sam.Waugh@cs.utas.edu.au})  
                    Department of Computer Science, University of Tasmania  
                    GPO Box 252C, Hobart, Tasmania 7001, Australia
                    \item[c)]\textbf{Date received:} December 1995
                \end{itemize}
            \item \textbf{Attribute Information:}  
        Given is the attribute name, attribute type, the measurement unit, and a brief description.
        \end{enumerate}

        \vspace{5pt}
        \renewcommand{\arraystretch}{1.2}
        \begin{center}
        \begin{tabular}{l l c l}
            \toprule
            \textbf{Name} & \textbf{Data Type} & \textbf{Meas.} & \textbf{Description} \\
            \midrule
            \textbf{x: Rings}  & Integer    &  & +1.5 gives the age in years \\
            \textbf{y: Height} & Continuous & mm & with meat in shell \\
            \bottomrule
        \end{tabular}
        \end{center}
    \end{tcolorbox}
    \caption{Example ground-truth-free textual description ($D'_i$) for a Tübingen pair (here $i=8$). Formatting was added here to help the reader but raw text is given to the LLM.}
    \label{fig:text_description}
\end{figure}

Then, we parse the causal direction from the answer. We considered GPT-4o and GPT-4o-mini with default hyperparameters; due to stochasticity we varied the seed which was assigned values 0, 1, $\dots$, 19.

\textbf{L2D-CD model:} L2D-CD was implemented according to Section \ref{sec:training_procedure}, where for each Tübignen pair~$i$, the variables remain constant as $(u_i,v_i) = (\texttt{First column (x)},\texttt{Second column (y)})$, the numerical variables are the two columns $(X_u, X_v) = (X_1, X_2)$, and the context $C_i$ is the ground-truth-free text description $D'_i$. For simplicity, we exclude numerical variables $X_u, X_v$ from the input to the deferral function, so with constant $(u,v)$'s across Tübingen pairs our deferral function becomes:
\begin{align*}
    r(x_i) = \text{Classifier}\left( \text{Embedding}(D'_i)\right)
\end{align*}
where ``Embedding'' refers to a model that converts the text description to a numerical vector embedding, and ``Classifier'' refers to any classifier from the space of this embedding. In practice, we observe small training samples (from 11 to 36) after restriction to the set $S$ as in Section \ref{sec:training_procedure}, motivating us to use random forests \citep{breiman2001rf} as classifiers; we resorted to the scikit-learn implementation \citep{pedregosa2011slmlip}. For the embedding,  we use OpenAI's \texttt{text-embedding-3-small} model and perform dimensionality reduction according to the recommended practice on the \href{https://platform.openai.com/docs/guides/embeddings}{OpenAI website}. To select hyperparameters for the L2D method, we perform hyperparameter search on a grid of the following hyperparameters : 10, 50, 100 for the random forest's \texttt{n\_estimators}; 2, 5 for its \texttt{min\_samples\_split};  5, 10, 15, 20, 50 for the size of the embedding.

After evaluating these hyperparameters across all possible pairs of experts and CD methods under consideration, and across 20 random training seeds, we select \texttt{n\_estimators} = 100 and \texttt{min\_samples\_split} = 5 for the random forest, and 50 dimensions for the embedding model. Evaluation is done using leave-one-out (LOO) cross-validation on the training set, using the loss from Equation~\ref{eq:deferral-func-org} and the training procedure from \ref{sec:training_procedure}. As a metric for selection, LOO sample-wise losses are averaged for each expert, CD method and random training seed, then for each of the three aforementioned types of experts, then across these three types; this is done in order to balance performance across such types. L2D-CD with the selected hyperparameters is then retrained for each expert, CD method and random seed using the full training set.

\textbf{Baselines}: In order to assess whether the description-wise heterogeneity of the L2D deferral function~$r$ 
impacts performance compared to randomly deferring to either method,
we introduce a baseline.
In this baseline, the probability of deferring to the expert is set as the fraction of correct expert predictions 
on the same set $S$ used to train L2D-CD's classifier. 
The causal direction of the testing set is then predicted as a Bernoulli of this probability. Given that baselines randomly sample causal predictions, in contrast to L2D predictions which are deterministic, we consider 20 random seeds for the sampling.

\textbf{L2D-CD consistently improves accuracy compared to the expert and CD alone}:  For each pair of expert and CD method, Table \ref{tab:acc} presents the testing accuracies for the CD method alone, for the expert alone, and for the L2D-CD and baseline combinations of these two methods. L2D-CD almost always improves over all other methods, showing its capability to combine accurate predictions from the expert and the CD method. Notably, it outperforms both synthetic experts and real-world LLM-based ones. While Tübingen pairs are known to have been memorized by LLMs since they are available online \citep{kiciman2023crallmoanffc}, L2D-CD still generally improves on the LLM experts, regardless of whether memorization occurs or not. On the other hand, the baseline's average accuracy interpolates between those of the CD method and the expert, showing the necessity of instance-dependent predictions.

\begin{table}

\caption{Hold-out accuracies by possible combinations of causal discovery (CD) method and expert, averaged by random seeds w.r.t. stochastic expert predictions, for random forest training and baseline sampling predictions (20 seeds for each). Associated standard errors are shown after ``$\pm$''; Notably, the standard error is always zero for CD methods and for $p$-experts, as they are deterministic.}
\begin{center}
\begin{tabular}{ll|cc>{\columncolor{blue!5}}c >{\color{gray}} c}
\toprule
 CD & Expert & CD Acc & Expert Acc &  L2D-CD Acc & Baseline Acc\\
\midrule
\multirow[c]{10}{*}{LiNGAM} & EMP & $0.442\pm0.000$ & $0.538\pm0.000$ & $\boldsymbol{0.661\pm0.005}$ & $0.489\pm0.011$ \\
 & CMP & $0.442\pm0.000$ & $0.635\pm0.000$ & $\boldsymbol{0.700\pm0.004}$ & $0.554\pm0.010$ \\
 & CEP & $0.442\pm0.000$ & $\boldsymbol{0.692\pm0.000}$ & $0.691\pm0.006$ & $0.603\pm0.007$ \\
 & CEM & $0.442\pm0.000$ & $0.673\pm0.000$ & $\boldsymbol{0.749\pm0.005}$ & $0.595\pm0.009$ \\
 & BMP & $0.442\pm0.000$ & $0.481\pm0.000$ & $\boldsymbol{0.648\pm0.006}$ & $0.466\pm0.013$ \\
 & BEP & $0.442\pm0.000$ & $0.538\pm0.000$ & $\boldsymbol{0.681\pm0.004}$ & $0.486\pm0.009$ \\
 & BEM & $0.442\pm0.000$ & $0.519\pm0.000$ & $\boldsymbol{0.626\pm0.007}$ & $0.476\pm0.012$ \\
 & BCP & $0.442\pm0.000$ & $0.635\pm0.000$ & $\boldsymbol{0.730\pm0.004}$ & $0.540\pm0.008$ \\
 & BCM & $0.442\pm0.000$ & $0.615\pm0.000$ & $\boldsymbol{0.687\pm0.003}$ & $0.532\pm0.009$ \\
 & BCE & $0.442\pm0.000$ & $0.673\pm0.000$ & $\boldsymbol{0.731\pm0.004}$ & $0.568\pm0.009$ \\
\midrule
\multirow[c]{10}{*}{bQCD} & EMP & $0.692\pm0.000$ & $0.538\pm0.000$ & $\boldsymbol{0.788\pm0.005}$ & $0.623\pm0.009$ \\
 & CMP & $0.692\pm0.000$ & $0.635\pm0.000$ & $\boldsymbol{0.854\pm0.003}$ & $0.656\pm0.012$ \\
 & CEP & $0.692\pm0.000$ & $0.692\pm0.000$ & $\boldsymbol{0.857\pm0.005}$ & $0.688\pm0.013$ \\
 & CEM & $0.692\pm0.000$ & $0.673\pm0.000$ & $\boldsymbol{0.779\pm0.002}$ & $0.683\pm0.009$ \\
 & BMP & $0.692\pm0.000$ & $0.481\pm0.000$ & $\boldsymbol{0.775\pm0.005}$ & $0.620\pm0.011$ \\
 & BEP & $0.692\pm0.000$ & $0.538\pm0.000$ & $\boldsymbol{0.796\pm0.005}$ & $0.633\pm0.010$ \\
 & BEM & $0.692\pm0.000$ & $0.519\pm0.000$ & $\boldsymbol{0.734\pm0.005}$ & $0.618\pm0.006$ \\
 & BCP & $0.692\pm0.000$ & $0.635\pm0.000$ & $\boldsymbol{0.856\pm0.003}$ & $0.668\pm0.012$ \\
 & BCM & $0.692\pm0.000$ & $0.615\pm0.000$ & $\boldsymbol{0.722\pm0.004}$ & $0.657\pm0.009$ \\
 & BCE & $0.692\pm0.000$ & $0.673\pm0.000$ & $\boldsymbol{0.753\pm0.002}$ & $0.676\pm0.011$ \\
\midrule
\multirow[c]{10}{*}{RECI} & EMP & $0.654\pm0.000$ & $0.538\pm0.000$ & $\boldsymbol{0.772\pm0.004}$ & $0.590\pm0.009$ \\
 & CMP & $0.654\pm0.000$ & $0.635\pm0.000$ & $\boldsymbol{0.740\pm0.003}$ & $0.652\pm0.010$ \\
 & CEP & $0.654\pm0.000$ & $0.692\pm0.000$ & $\boldsymbol{0.846\pm0.004}$ & $0.684\pm0.010$ \\
 & CEM & $0.654\pm0.000$ & $0.673\pm0.000$ & $\boldsymbol{0.785\pm0.004}$ & $0.682\pm0.008$ \\
 & BMP & $0.654\pm0.000$ & $0.481\pm0.000$ & $\boldsymbol{0.768\pm0.004}$ & $0.585\pm0.011$ \\
 & BEP & $0.654\pm0.000$ & $0.538\pm0.000$ & $\boldsymbol{0.883\pm0.003}$ & $0.593\pm0.008$ \\
 & BEM & $0.654\pm0.000$ & $0.519\pm0.000$ & $\boldsymbol{0.776\pm0.002}$ & $0.588\pm0.009$ \\
 & BCP & $0.654\pm0.000$ & $0.635\pm0.000$ & $\boldsymbol{0.824\pm0.004}$ & $0.652\pm0.010$ \\
 & BCM & $0.654\pm0.000$ & $0.615\pm0.000$ & $\boldsymbol{0.767\pm0.006}$ & $0.642\pm0.007$ \\
 & BCE & $0.654\pm0.000$ & $0.673\pm0.000$ & $\boldsymbol{0.844\pm0.003}$ & $0.663\pm0.010$ \\
\midrule
\multirow[c]{3}{*}{LiNGAM} & $\epsilon = 0.05$ & $0.442\pm0.000$ & $0.535\pm0.001$ & $\boldsymbol{0.676\pm0.001}$ & $0.484\pm0.002$ \\
 & $\epsilon = 0.1$ & $0.442\pm0.000$ & $0.531\pm0.001$ & $\boldsymbol{0.672\pm0.002}$ & $0.483\pm0.002$ \\
 & $\epsilon = 0.2$ & $0.442\pm0.000$ & $0.512\pm0.002$ & $\boldsymbol{0.648\pm0.002}$ & $0.476\pm0.002$ \\
\midrule
\multirow[c]{3}{*}{bQCD} & $\epsilon = 0.05$ & $0.692\pm0.000$ & $0.535\pm0.001$ & $\boldsymbol{0.795\pm0.001}$ & $0.632\pm0.002$ \\
 & $\epsilon = 0.1$ & $0.692\pm0.000$ & $0.531\pm0.001$ & $\boldsymbol{0.793\pm0.001}$ & $0.630\pm0.002$ \\
 & $\epsilon = 0.2$ & $0.692\pm0.000$ & $0.512\pm0.002$ & $\boldsymbol{0.785\pm0.001}$ & $0.625\pm0.002$ \\
\midrule
\multirow[c]{3}{*}{RECI} & $\epsilon = 0.05$ & $0.654\pm0.000$ & $0.535\pm0.001$ & $\boldsymbol{0.873\pm0.002}$ & $0.593\pm0.002$ \\
 & $\epsilon = 0.1$ & $0.654\pm0.000$ & $0.531\pm0.001$ & $\boldsymbol{0.863\pm0.003}$ & $0.593\pm0.002$ \\
 & $\epsilon = 0.2$ & $0.654\pm0.000$ & $0.512\pm0.002$ & $\boldsymbol{0.815\pm0.005}$ & $0.592\pm0.002$ \\
\midrule
\multirow[c]{2}{*}{LiNGAM} & GPT4o & $0.442\pm0.000$ & $\boldsymbol{0.751\pm0.001}$ & $0.747\pm0.001$ & $0.662\pm0.002$ \\
 & GPT4o-mini & $0.442\pm0.000$ & $0.755\pm0.001$ & $\boldsymbol{0.771\pm0.001}$ & $0.669\pm0.002$ \\
\midrule
\multirow[c]{2}{*}{bQCD} & GPT4o & $0.692\pm0.000$ & $0.751\pm0.001$ & $\boldsymbol{0.795\pm0.002}$ & $0.739\pm0.003$ \\
 & GPT4o-mini & $0.692\pm0.000$ & $0.755\pm0.001$ & $\boldsymbol{0.820\pm0.001}$ & $0.746\pm0.003$ \\
\midrule
\multirow[c]{2}{*}{RECI} & GPT4o & $0.654\pm0.000$ & $0.751\pm0.001$ & $\boldsymbol{0.773\pm0.002}$ & $0.733\pm0.002$ \\
 & GPT4o-mini & $0.654\pm0.000$ & $0.755\pm0.001$ & $\boldsymbol{0.795\pm0.001}$ & $0.742\pm0.002$ \\
\bottomrule
\end{tabular}
\label{tab:acc}
\end{center}
\end{table}

\textbf{L2D-CD can identify strong and weak domains for the expert}: While the accuracy results indicate improved performance of L2D-CD on all pairs, one might wonder whether L2D-CD can identify domains where the expert is strong and those where the expert is weak. To assess this, we focus on synthetic $\epsilon$-experts and $p$-experts where performance is controlled through domain-wise probabilities $p_d$. Let $r(x; \text{CD}, \text{Expert}, \epsilon')$ be the deferral function corresponding to a given causal discovery method $\text{CD}$, a given expert $\text{Expert}$ and a random variable $\epsilon'$ capturing sources of uncertainty, such as the expert seed for stochastic experts, the training seed for random forests, or the sampling seed for baselines. Then, the probability that $r$ chooses $\text{Expert}$ on domain $d$ can be computed as
\begin{align*}
p(\text{Expert chosen by r} | \text{Expert}, d) = \mathbb{E}_{X, \epsilon', \text{CD}}[1_{\{r(X; \text{CD}, \text{Expert}, \epsilon') = \text{Expert}\}} \ | \ X \in d]
\end{align*}
where we average over the CD method, the samples in the domain and uncertainty from the expert and the deferral function fitting. We say that $r$ is \textbf{domain-consistent} for \text{Expert} if, on average, it defers to Expert more frequently on any domain $d_+$ where \text{Expert} is strong than any domain $d_-$ where \text{Expert} is weak:
\begin{align}
H^1_{d_+,d_-,r,\text{Expert}} : p(\text{Expert chosen by r} | \text{Expert}, d_+) > p(\text{Expert chosen by r} | \text{Expert}, d_-)  \label{eq:pairwise_consistency}
\end{align}
Notably, this would enable setting a threshold separating strong and weak domains for the expert. For experts with domain-wise probabilities $p_d$, strong domains are defined as $p_d > 0.5$ and weak domains $p_d < 0.5$. We exclude LLMs for analysis as (i) they do not have such ground-truth probabilities, and (ii) their domain-wise accuracies are all above 0.5, thus are strong as in the previous definition on all domains. We describe our approach to evaluate domain consistency using hypothesis testing in Appendix~\ref{supp:domain_consistency}. Table~\ref{tab:domain_consistency} presents whether domain consistency holds depending on whether $r$ is L2D-CD or the baseline and on the synthetic expert. Notably, we can see that the L2D-CD is domain-consistent for almost every synthetic expert, while the baseline is never domain-consistent for any synthetic expert. This shows that L2D-CD can capture the strengths and weaknesses of experts, while deferral based on a constant probability (unsurprisingly) cannot.

\begin{table}[t]
\caption{Domain consistency of each combination method for each synthetic expert}
\label{tab:domain_consistency}
\centering
\begin{tabular}{lccccccc}
\toprule
 & $\epsilon = 0.05$ & $\epsilon = 0.1$ & $\epsilon = 0.2$ & BCE & BCM & BCP & BEM  \\
\midrule
L2D-CD & Yes & Yes & Yes & Yes & Yes & Yes & Yes \\
Baseline    & No  & No  & No  & No  & No  & No  & No  \\
\bottomrule
\end{tabular}

\quad

\begin{tabular}{lcccccc}
\toprule
 & BEP &BMP & CEM & CEP & CMP & EMP \\
\midrule
L2D-CD & Yes & No  & Yes & Yes & Yes & Yes \\
Baseline    & No & No  & No  & No  & No  & No  \\
\bottomrule
\end{tabular}
\end{table}

\section{Extension to graphs with 3 variables or more}

To extend L2D-CD to graphs of 3 variables or more, we propose building on methods for ranking from pairwise comparisons \citep{braverman2007nswr, ailon2011aalafrfppwaaoqc, rajkumar2014ascpoafrafpd, mao2017mraeafns, falahatgar2018tlomrapl, ren2021oscualbferfnc}, where a ranking over $V$ is a function $\pi$ that is bijective from $V$ to $\{1, \dots, |V|\}$.
We let $\pi(u) < \pi(u')$ indicate that $u$ is ranked before $u'$ for any $u,u' \in V$.
While these methods differ in their exact problem formulation, they all amount to learning a ranking $\hat{\pi}(u)$ over items $u$ in a set $V$ from samples $(u_i, u_i', y_i)$, where $y_i \in \{-1, 1\}$ indicates a comparison between $u_i$ and $u'_i$, with  $y_i = 1$ indicating that $u_i$ is deemed as ranked before $u'_i$ and $y_i = -1$ as the converse. Thus, the learning algorithm $\mathcal{A}$ attempts to find a ranking $\pi$ that best fits the individual comparisons; the latters are allowed to be contradictory. Thus we propose generalizing our method to causal discovery on more than two variables as follows.

\textbf{Graph notations:} Let $G$ be a graph having nodes $V_G$, where every node is represented by its name, and edges $E_G$. Define $\Sigma(V_G,E_G)$ as a matrix of pairwise ancestries derived from $E_G$, where $(u,u')$ is equal to $1$ if $u$ precedes $u'$ in $E_G$, $-1$ if $u'$ prcedes $u$, and $0$ if no ancestry relationship between $u$ and $u'$ exists. Additionally, let $C_G$ be the graph's textual context, and $X_G = (X_u)_{u \in V_G}$ its numerical data.

\textbf{Problem definition:}  Let $u,v$ be the textual names of nodes, $X$ the full numerical data of a graph, and $C$ a textual context. Assume access to an expert $\text{Expert}(C,u,v)$ and a causal discovery oracle $\text{CD}(u,v;X)$, both of which determine the ancestry relationship between $u$ and $v$.
The causal discovery oracle applies the causal discovery method to 
$X$ and, like the expert, may return the absence of ancestry.

\textbf{Training:} Assume that we have a training set of graphs $\mathcal{G}_\text{train} = (V_{G_i}, E_{G_i}, C_{G_i}, X_{G_i})_{i=1,\dots,m_{\text{train}}}$. Then, train a deferral function $r(C,u,v,X)$ between the expert and the causal discovery oracle using L2D-CD where the training sample is composed of each pair of nodes in each graph and its ancestry status, i.e. the training set is formed by $(C_{G_i}, u, u', X_{G_i})_{u, u' \in V_{G_i}, i=1,\dots,m_{\text{train}}}$ as features, and $\left(\Sigma_{u,u'}(V_{G_i},E_{G_i})\right)_{u, u' \in V_{G_i}, i=1,\dots,m_{\text{train}}}$ as the corresponding labels.

\textbf{Inference:} For every hold-out graph $G$, where we have access to $V_G, C_G, X_G$, sample pairs of nodes $u,u' \in V_G$, infer their ancestry relationship $\hat{y}_{u,u'}$ using L2D-CD, decide on how to handle $y_{u,u'} = 0$, e.g. discard; apply $\mathcal{A}$ to the resulting comparisons to obtain a topological ordering $\hat{\pi}$; optionally deduce an edge set $\hat{E}_G$ using an edge pruning method \citep{buhlmann2014ccamhdosapr}.
    
Algorithms $\mathcal{A}$ for ranking from pairwise preferences typically benefit from convergence to a ground-truth or optimal ranking in a moderately scaling number of steps. 
Key challenges will be  establishing convergence to any of the multiple topological orderings allowed by a causal graph depending on the accuracy of the deferral function, and handling the absence of ancestry in pairwise comparisons.

\section{Conclusion}

We have designed a procedure leveraging learning-to-defer to combine two pairwise causal discovery methods, one conventional method and one expert. Experiments on Tübingen pairs showed that the combined method generally improves over each separate method, and so for both synthetic and real-world LLM-based experts. The learnt deferral function can also identify the expert's strong and weak domains. Note that our methodology can also be applied to human knowledge. An inherent limitation of the L2D-CD approach is the need for a training set, while an improvable one is that our training procedure does not generalize straightforwardly to more than two methods.%
We also did not yet implement our strategy to generalize L2D-CD for bivariate causal discovery to more general causal discovery, which is future work.

\subsubsection*{Acknowledgements}

We sincerely thank Philippe Brouillard for helpful comments on the manuscript. O.C. acknowledges support from the EPSRC Centre for Doctoral Training in Modern Statistics and Statistical Machine Learning (EP/S023151/1) and Novo Nordisk for doctoral studies, as well as from Mitacs for the internship at ServiceNow Research. I.M. acknowledges support by an NSERC Discovery grant (RGPIN-2019-06512), and a Canada CIFAR AI chair. D.M. acknowledges support via FRQNT doctoral training scholarship for his graduate studies.

\newpage

\bibliography{iclr2025_conference}
\bibliographystyle{iclr2025_conference}

\clearpage
\appendix

\section{Proof of Lemma \ref{lem:bound}
\label{supp:proof}}

The result follows from following the proof of Theorem 6 from \citet{mao2024tsltdwme}, except that we change the upper bound of $\mathcal{C}_{L_\text{def}}(h,r,x) - \inf_{r \in \mathcal{R}} \mathcal{C}_{L_\text{def}}(h,r,x)$, where for any loss $\ell(g,x,y)$ where $g$ is a predictor, $\mathcal{C}_{\ell}(g,x) := \mathbb{E}_{y|x} \left[\ell(g,x,y)\right]$. We aim to prove that
\begin{align*}
    \mathcal{C}_{L_\text{def}}(h,r,x) - \inf_{r \in \mathcal{R}} \mathcal{C}_{L_\text{def}}(h,r,x) \leq     \begin{cases}
      \Gamma_2\left(\mathcal{C}_{L_h}(r,x) - \inf_{r \in \mathcal{R}} \mathcal{C}_{L_h}(r,x)\right) & \text{if $\Gamma_2$ is linear}\\
      n_e\Gamma_2(\mathcal{C}_{L_h}(r,x) - \inf_{r \in \mathcal{R}} \mathcal{C}_{L_h}(r,x)) & \text{otherwise}\\
    \end{cases}
\end{align*}

Let $\mathcal{X}_\circ := \{x \ \ \ | \ \ \ h(x) = h_j(x) \ \ \forall j = 1, \dots, n_e\}$ and $x \in \mathcal{X}$. If $x \in \mathcal{X}_\circ$ then it turns out that the bound holds since the LHS is zero, as $L_\text{def}(h,r,x,y) = \mathbb{I}_{h(x) \neq y}$ does not depend on $r$, and the RHS is non-negative. Now assume $x \notin \mathcal{X}_\circ$. Then, with the convention that $h_0 = h$ and $c_0(x,y) = \mathbb{I}_{h(x) \neq y}$, for any $y$, $\sum_{j=0}^{n_e} \bar{c}_j(x,y) \geq 1$, as each each $\bar{c}_j(x,y)$ is binary and, from $y$ being binary, all $\bar{c}_j(x,y)$'s being zero would imply that $\forall j = 0, \dots, n_e, \ \ h_j(x) = 1 - y$, which contradicts $x \notin \mathcal{X}_\circ$. Similarly, $\sum_{j=0}^{n_e} \bar{c}_j(x,y) \leq n_e$ as at least one $\bar{c}_j(x,y)$ should be zero (this does not require $y$ being binary). Then, we can repeat the steps of the original proof of \citet{mao2024tsltdwme} to obtain the upper-bound, using $1 \leq \mathbb{E}_{y|x}\left[\sum_{j=0}^{n_e} \bar{c}_j(x,y)\right] \leq n_e$. This completes the general consistency bound.

For the part where $\mathcal{H}$ is a singleton $h_0$, this follows from the argument of the $\Gamma_1$ function being zero as a result of the singleton.

\section{Tübingen pairs by domain and training or testing set}
\label{app:pairs}

\begin{table}[h]    
\caption{Tübingen pairs, denoted by their numerical identifiers, for each domain and training/testing set.}
    \centering
\begin{tabular}{lcc}
\toprule
 & Training set & Testing set \\
\midrule
Biology & 7, 9, 70, 78, 79, 90, 92 & 5, 6, 8, 10, 11, 80, 89, 91 \\
\midrule

Climate/Environment & \makecell{ 1, 3, 4, 13, 15, 19, 21, 42, 48,\\ 50, 72, 77, 82, 83, 94, 95} & \makecell{ 2, 14, 16, 20, 43, 44, 45, 46,\\ 49, 51, 69, 73, 81, 87, 93, 96} \\
\midrule

Economics/Finance & \makecell{12, 47, 57, 58, 60, 61, \\62, 63, 67, 68, 86} & \makecell{17, 56, 59, 64, 65, 66,\\ 74, 75, 76, 84, 99} \\
\midrule
Medicine & 18, 22, 34, 36, 39, 40, 88, 107 & 23, 24, 33, 35, 37, 38, 41, 85 \\
\midrule

Physics & 26, 28, 30, 31, 32, 97, 103, 104 & \makecell{25, 27, 29, 98, 100,\\ 101, 102, 106, 108} \\
\bottomrule
\end{tabular}
\label{tab:pairs}
\end{table}

\section{Hypothesis testing to assess domain consistency}
\label{supp:domain_consistency}
To take account of uncertainty, for fixed $r$ and Expert, we assessed Equation \ref{eq:pairwise_consistency} by performing a statistical test with $H^1_{d_+,d_-,r,\text{Expert}}$ as the alternative hypothesis and
\begin{align*}
    H^0_{d_+,d_-,r,\text{Expert}} : p(\text{Expert chosen by r} | \text{Expert}, d_+) \leq p(\text{Expert chosen by r} | \text{Expert}, d_-)
\end{align*}
as the null hypothesis for each strong/weak domain pair $(d_+, d_-)$. This was done using Fisher's exact test over the sets of binary values $I_{d^+}$ and $I_{d^-}$ where 
\begin{align*}
    I_d := (1_{\{r(x_i; \text{CD}, \text{Expert}, \epsilon') = \text{Expert}\}})_{i \in \mathcal{T}_d, \text{CD}, \epsilon'},
\end{align*}
with $\mathcal{T}_d$ denoting the intersection of the testing set and the domain $d$. This yields a p-value $\text{pval}(d_+,d_-,r,\text{Expert})$. Then, we assess domain consistency of $r$ for Expert by computing a p-value $\text{pval}(r,\text{Expert})$ for the null hypothesis $H^0_{r,\text{Expert}} = \bigcup_{d_+,d_-,} H^0_{d_+,d_-,r,\text{Expert}}$ and the alternative hypothesis $H^1_{r,\text{Expert}} = \bigcap _{d_+,d_-,} H^1_{d_+,d_-,r,\text{Expert}}$. This is a classical instance of an intersection-union test \citep{berger1996btiutaecs} and for its p-value we can take $\text{pval}(r,\text{Expert}) = \max_{d_+, d_-} \text{pval}(d_+,d_-,r,\text{Expert})$ \citep{schuirmann1987acottostpatpafateoab}. In the end, domain consistency of $r$ for Expert is defined as $\text{pval}_\text{corrected}(r,\text{Expert}) < 0.05$, where we adjust all p-values $\text{pval}(r,\text{Expert})$ jointly using Benjamini-Hochberg correction \citep{benjamini1995ctfdrapapatmt} to obtain  $\text{pval}_\text{corrected}(r,\text{Expert})$.

\end{document}